\documentclass{article}

     \usepackage[preprint]{neurips_2019}

\usepackage{wrapfig}
\usepackage{microtype}
\usepackage{graphicx}
\usepackage{booktabs} 
\usepackage{stfloats}
\usepackage[utf8]{inputenc}
\usepackage[french,english]{babel}
\usepackage{cancel}
\usepackage{amsmath}
\usepackage{amssymb}
\usepackage{amsthm}
\usepackage{algorithm}
\usepackage{algorithmic}
\usepackage{varwidth}
\usepackage{subcaption}
\usepackage{comment}

\usepackage{amsmath,amsfonts,bm}

\def\eqref#1{equation~\ref{#1}}

\def\1{\bm{1}}

\def\vtheta{{\bm{\theta}}}

\def\vw{{\bm{w}}}
\def\vx{{\bm{x}}}

\def\vz{{\bm{z}}}

\DeclareMathAlphabet{\mathsfit}{\encodingdefault}{\sfdefault}{m}{sl}
\SetMathAlphabet{\mathsfit}{bold}{\encodingdefault}{\sfdefault}{bx}{n}

\newcommand{\E}{\mathbb{E}}

\newcommand{\R}{\mathbb{R}}

\newcommand{\x}{{\mathbf x}}
\newcommand{\z}{{\mathbf z}}

\usepackage[utf8]{inputenc} 
\usepackage[T1]{fontenc}    
\usepackage{hyperref}       
\usepackage{url}            
\usepackage{booktabs}       
\usepackage{amsfonts}       
\usepackage{nicefrac}       
\usepackage{microtype}      
\usepackage{amsmath}
\usepackage{amssymb}
\usepackage{algorithm}
\usepackage{algorithmic}
\usepackage{hyperref}
\usepackage{tabularx}
\usepackage{graphicx}
\usepackage[title]{appendix}

\newcommand{\norm}[1]{\left\lVert#1\right\rVert}

\title{Maximum Entropy Generators \\ for Energy-Based Models}

\author{
Rithesh Kumar$^1$\thanks{Correspondence to {\tt rithesh.kumar@umontreal.ca}}
 \qquad Sherjil Ozair$^1$ \qquad
Anirudh Goyal$^1$ \\
\textbf{Aaron Courville}$^{1, 2}$ \qquad \textbf{Yoshua Bengio}$^{1, 2, 3}$
\\
\\
$^1$Mila, Université de Montréal \\$^2$Canadian Institute for Advanced Research (CIFAR) \\$^3$ Institute for Data Valorization (IVADO)
}

\begin{document}

\maketitle

\begin{abstract}
Maximum likelihood estimation of energy-based models is a challenging problem due to the intractability of the log-likelihood gradient. In this work, we propose learning both the energy function and an amortized approximate sampling mechanism using a neural generator network, which provides an efficient approximation of the log-likelihood gradient. The resulting objective requires maximizing entropy of the generated samples, which we perform using recently proposed nonparametric mutual information estimators. Finally, to stabilize the resulting adversarial game, we use a zero-centered gradient penalty derived as a necessary condition from the score matching literature.
The proposed technique can generate sharp images with Inception and FID scores competitive with recent GAN techniques, does not suffer from mode collapse, and is competitive with state-of-the-art anomaly detection techniques.




\end{abstract}

\section{Introduction}

Unsupervised learning promises to take advantage from unlabelled data, and is regarded as crucial for artificial intelligence \citep{lake2017building}. Energy-based modeling (EBMs, \citet{lecun2006}) is a family of unsupervised learning methods focused on learning an energy function, i.e.,  an unnormalized log density of the data. This removes the need to make parametric assumptions about the data distribution to make the normalizing constant ($Z$) tractable.  However, in practice, due to the very same lack of restrictions, learning high-quality energy-based models is fraught with challenges. To avoid explicitly computing $Z$ or its gradient, 
 Contrastive Divergence~\citep{Hinton-PoE-2000} and Stochastic Maximum Likelihood~\citep{Younes98,Tieleman08} rely on Markov Chain Monte Carlo (MCMC) to approximately sample from the energy-based model. However, MCMC-based sampling approaches frequently suffer from long mixing times for high-dimensional data. Thus, training of energy-based models has not remained competitive with other unsupervised learning techniques such as variational auto-encoders~\citep{Kingma+Welling-ICLR2014} and generative adversarial networks or GANs~\citep{Goodfellow-et-al-NIPS2014}.


In this work, we propose Maximum Entropy Generators (MEG), a framework in which we train both an energy function and an approximate sampler, which can either be fast (using a generator network $G$) or uses $G$ to initialize a Markov chain in the latent space of the generator. Training such a generator properly requires entropy maximization of the generator's output distribution, for which we take advantage of recent advances in nonparametric mutual information maximization \citep{belghazi2018mine,Hjelm-et-al-DIM-2018,oord2018representation,poole18variational}.

To evaluate the efficacy of the proposed technique, we compare against other state-of-the-art techniques on image generation, accurate mode representation, and anomaly detection. We demonstrate that the proposed technique is able to generate CIFAR-10 samples which are competitive with WGAN-GP \citep{gulrajani2017improved} according to the Fréchet Inception Distance \citep{heusel2017gans} and Inception Score \citep{salimans2016improved}, and is able to generate samples of all the $10^4$ modes of 4-StackedMNIST at the correct data frequencies.

We demonstrate that our technique trains energy functions useful for anomaly detection on the KDD99 dataset, and that it performs as well as state-of-the-art anomaly detection techniques which were specially designed for the task, and vastly outperform other energy-based and generative models for anomaly detection.




To summarize our contributions, we propose maximum entropy generators (MEG), a novel framework for training energy-based models using amortized neural generators and mutual information maximization. We show that the resulting energy function can be successfully used for anomaly detection, and outperforms recently published results with energy-based models. We show that MEG generates sharp images -- with competitive Inception and FID scores -- and accurately captures more modes than standard GANs, while not suffering from the common mode-mixing issue of many maximum likelihood generative models which results in blurry samples.

\begin{figure}[t]
    \centering
    \includegraphics[scale=.8]{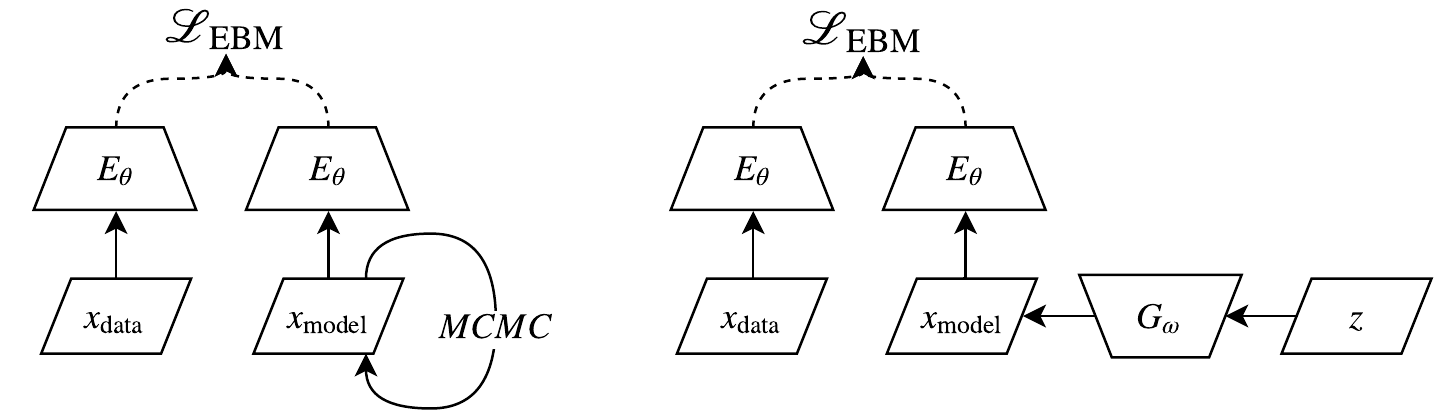}
    \caption{\textbf{Left}: Traditional maximum likelihood training of energy-based models. \textbf{Right:} Training of maximum entropy generators for energy-based models}
    \label{fig:ebm_comparison}
\end{figure}

\section{Background}
\label{sec:negative-phase-gradient}

Let $\vx$ denote a sample in the data space $\cal X$ and $E_\theta:{\cal X} \rightarrow \R$ an energy function corresponding to the negative logarithm of an unnormalized estimated
density density function 
\begin{align}
    p_\theta(\vx) = \frac{e^{-E_\theta(\vx)}}{Z_\theta} \propto e^{-E_\theta(\vx)}
\end{align}
where 
   $Z_\theta := \int e^{-E_\vtheta(\vx)} d\vx$
is the normalizing constant or partition function.
%
Let $p_D$ be the training distribution, from which the training set is drawn. Towards optimizing the parameters $\theta$ of the energy function,
the maximum likelihood parameter gradient is
\begin{align}
\label{eq:ll-gradient}
  \frac{\partial \E_{\vx \sim p_D} [- \log p_\theta(\vx)]}{\partial \theta}  =
  \E_{\vx \sim p_D} \left[ \frac{\partial E_\theta(\vx)}{\partial \theta} \right] - \E_{\vx \sim p_\theta(\vx)} \left[ \frac{\partial E_\theta(\vx)}{\partial \theta} \right]
\end{align}
where the second term is the gradient of $\log Z_\theta$, and the sum of the two expectations is zero when training has converged, with expected energy gradients in the positive phase (under the data $p_D$) matching those under the negative phase (under $p_\theta(\vx)$). Training thus consists in trying to separate two distributions: the positive phase distribution (associated with the data) and the negative phase distribution (where the model is free-running and generating configurations by itself). This observation has motivated the pre-GAN idea presented by~\citet{Bengio-2009-book} that ``model samples are negative examples" and a classifier could be used to learn an energy function if it separated the data distribution from the model's own samples. Shortly after introducing GANs, \citet{Goodfellow2014}  also made a similar connection, related to noise-contrastive estimation~\citep{Gutmann+Hyvarinen-2010}. One should also recognize the similarity between Eq.~\ref{eq:ll-gradient} and the objective function for Wasserstein GANs or WGAN~\citep{Arjowski-et-al-2017}.

The main challenge in Eq.~\ref{eq:ll-gradient} is to obtain samples from the distribution $p_\theta$ associated with the energy function $E_\theta$. Although having an energy function is convenient to obtain a score allowing comparison of the relative probability for different $\vx$'s, it is difficult to convert an energy function into a generative process. The commonly studied approaches for this are based on Markov Chain Monte Carlo, in which one iteratively updates a candidate configuration, until these configurations converge in  distribution to the desired distribution $p_\theta$. For the RBM, the most commonly used algorithms have been Contrastive Divergence~\citep{Hinton-PoE-2000} and Stochastic Maximum Likelihood~\citep{Younes98,Tieleman08}, relying on the particular structure of the RBM to perform Gibbs sampling. Although these MCMC-based methods are appealing, RBMs (and their deeper form, the deep Boltzmann machine) have not been competitive in recent years compared to autoregressive models~\citep{Oord-et-al-2016}, variational auto-encoders~\citep{Kingma+Welling-ICLR2014} and generative adversarial networks or GANs~\citep{Goodfellow-et-al-NIPS2014}. 

What has been hypothesized as a reason for poorer results obtained with energy-based models trained with an MCMC estimator for the negative phase gradient is that running a Markov chain in data space is fundamentally difficult when the distribution is concentrated (e.g, near manifolds) and has many modes separated by vast areas of low probability. This mixing challenge is discussed by ~\citet{Bengio-et-al-ICML2013} who argue that a Markov chain is very likely to produce only sequences of highly probable configurations: if two modes are far from each other and only local moves are possible (which is typically the case when performing MCMC), it becomes exponentially unlikely to traverse the ``desert'' of low probability that can separate two modes. This makes mixing between modes difficult in high-dimensional spaces with strong concentration of probability mass in some regions (e.g. corresponding to different categories) and very low probability elsewhere.



\section{Maximum Entropy Generators for Energy-Based Models}

We thus propose using an amortized neural sampler to perform fast approximate sampling to train the energy model. We begin by replacing the model distribution $p_\theta$ in in Eq.~\ref{eq:ll-gradient} by a neural generator $G$ parametrized by $\vw$. We define $P_G$ as the distribution of the outputs $G(z)$ for $\vz \sim p_z$ where $p_z$ is a simple prior distribution such as a standard Normal distribution.
\begin{align}
\label{regularizer}
   \frac{\partial \mathcal{L}_E}{\partial \theta}
    &= \E_{\vx \sim p_D} \left[ \frac{\partial E_\theta(\vx)}{\partial \theta} \right]
- \E_{\vx \sim p_G(\vx)} \left[ \frac{\partial E_\theta(\vx)}{\partial \theta} \right] 
\end{align}


To minimize the approximation error, $p_G$ must be close to $p_\theta$.
To do so, we tune $G$ to minimize the KL divergence $KL(p_G||p_\theta)$, which can be rewritten in terms of minimizing the energy of the samples from the generator while maximizing the entropy at the output of the generator:
\begin{align}
    KL(p_G||p_\theta) &= -H[p_G] - E_{p_G}[\log p_\theta(\vx)] \\
    &=  -H[p_G] + E_{p_G}[E_\theta(\vx)] + \log Z_{\theta}
\end{align}

When taking the gradient of $KL(p_G||p_\theta)$ with respect to the parameters $\vw$ of the generator, the log-partition function $\log Z_{\theta}$ disappears and we can optimize $\vw$ by minimizing
\begin{align}
\label{generator}
  \mathcal{L}_G = -H[p_G] + \E_{\vz \sim p_z} E_\theta(G(\vz))  
\end{align}
where $p_z$ is the prior distribution of the latent variable of the generator.

In order to approximately maximize the entropy $H[p_G]$ at the output of the generator, we use one recently proposed nonparametric mutual information maximization techniques \citep{belghazi2018mine,oord2018representation,Hjelm-et-al-DIM-2018}. \cite{poole18variational} show that these techniques can be unified into a single framework derived from the variational bound of \cite{Barber:2003:IAV:2981345.2981371}. Since the generator is deterministic, mutual information between inputs and outputs reduces to simply entropy of the outputs, since the conditional entropy of a deterministic function is zero:
\begin{align*}
I(X, Z) &= H(X) - H(X | Z) = H(G(Z)) - \cancelto{0}{H(G(Z) | Z)}
\end{align*}
In particular, we use the estimator from \cite{Hjelm-et-al-DIM-2018}, which estimates the Jensen-Shannon divergence between the joint distribution ($p(\vx, \vz)$) and the product of marginals ($p(\vx)p(\vz)$). We refer to this information measure as $I_{JSD}(X, Z)$. We found that the JSD-based estimator works better in practice than the KL-based estimator (which corresponds to the mutual information).

The estimator of \cite{Hjelm-et-al-DIM-2018} is given by
\begin{align}
    {\cal I}_{JSD}(X,Z) = \sup_{T \in \mathcal{T}} \E_{p(X,Z)}[-\mathrm{sp}(-T(X,Z))] - \E_{p(X)p(Z)}[\mathrm{sp}(T(X,Z))]
\end{align}
where $\mathrm{sp}(a)=\log(1+e^a)$ is the softplus function. The supremum is approximated using gradient descent on the parameters of the discriminator $T$.

With $X=G(Z)$ the output of the generator, ${\cal I}_{JSD}(G(Z),Z)$ is one of the terms to be maximized in the objective function for training $G$, which would maximize the generator's output entropy $H(G(Z))$.
\\
\\
Thus the final training objective to be minimized for the generator $G$ and the energy function $E$ is 
\begin{align}
  \mathcal{L}_G &= -{\cal I}_{JSD}(G(Z),Z)  + \E_{\vz \sim p_z} E_\theta(G(\vz)) \\
  \mathcal{L}_E &= \E_{\vx \sim p_{D}} E_\theta(\vx) - \E_{\vz \sim p_z} E_\theta(G(\vz))
  \label{eqn-final}
\end{align}
where $Z \sim p_z$, the latent prior (typically a $N(0,I)$ Gaussian).


\subsection{Improving training stability}

As can be seen from the above equations, the generator and the energy function are in an adversarial game, similar to generative adversarial networks \citep{Goodfellow-et-al-NIPS2014}. This makes optimization via simultaneous gradient descent challenging since the gradient vector field of such an optimization problem is non-conservative as noted by \cite{mescheder2017numerics}. This is particularly accentuated by the use of deep neural networks for the generator and the energy function.
In particular, we noticed that during training the magnitude of the energy function values would diverge.


To help alleviate this issue we look towards another technique for learning energy-based models called \emph{score matching} proposed by \cite{hyvarinen2005estimation}. Score matching estimates the energy function by matching the score functions of the data density and the model density, where the score function $\psi$ is the gradient of the log density with respect to the sample $\psi(\vx) = \frac{\partial \log p(\vx)}{\partial \vx}$. If $\psi_D(\vx)$ and $\psi_E(\vx)$ are the score functions under the data distribution and model distribution respectively, the score matching objective is given by
\begin{align}
\mathcal{J}_{SM} &= \mathbb{E}_{\vx \sim P_\text{D}} \left[ \norm{\psi_{D}(\vx) - \psi_{E}(\vx)}_2^2\right]. \nonumber
\end{align}
While the score function for the data distribution is typically unknown and would require estimation, Theorem 1 in \cite{hyvarinen2005estimation} shows that with partial integrations, the score matching objective can be reduced to the following objective which does not depend on the score function under the data distribution:
\begin{align}
    \mathcal{J}_{SM} &= \mathbb{E}_{\vx \sim P_\text{D}} \bigg[\sum_i \partial_i \psi_i(\vx) + \frac{1}{2} \psi_i(\vx)^2\bigg] \nonumber \\
    &= \mathbb{E}_{\vx \sim P_\text{D}} \bigg[\sum_i -\frac{\partial^2 E(\vx)}{\partial^2 \vx_i} + \frac{1}{2}\bigg(\frac{-\partial E(\vx)}{\partial \vx_i}\bigg)^2\bigg] \nonumber \\
    &= \mathbb{E}_{\vx \sim P_\text{D}} \bigg[\frac{1}{2} \norm{ \frac{\partial E(\vx)}{\partial \vx}}^{2}_{2}  + \sum_i -\frac{\partial^2 E(\vx)}{\partial^2 \vx_i}\bigg]
\end{align}

The above objective is hard to optimize when using deep neural networks because of the difficulty in estimating the gradient of the Hessian diagonal, so we use the first term in our objective, i.e. the zero-centered gradient penalty, pushing the data points to sit near critical points (generally a local minimum) of the energy function.

This term is also similar to the gradient penalty regularization proposed by \cite{gulrajani2017improved} which however is one-centered and applied on interpolations of the data and model samples, and is derived from the Lipschitz continuity requirements of Wasserstein GANs \citep{Arjowski-et-al-2017}.





\subsection{Improving sample quality via latent space MCMC}


Since MEG simultaneously trains a generator and a valid energy function, we can improve the quality of samples by biasing sampling towards high density regions. Furthermore, doing the MCMC walk in the latent space should be easier than in data space because the transformed data manifold (in latent space) is flatter than in the original observed data space, as initially discussed by~\citet{Bengio-et-al-ICML2013}. The motivation is also similar to that of the ``truncation trick'' used successfully by \citet{brock2018large}. However, we use an MCMC-based approach for this which is applicable to arbitrary latent distributions.

We use the Metropolis-adjusted Langevin algorithm (MALA, \citet{girolami2011riemann}), with Langevin dynamics producing a proposal distribution in the latent space as follows:
\\
\begin{align*}
\tilde{\z}_{t+1} = \vz_t - \alpha \frac{\partial E_\theta(G_\omega(\vz_t))}{\partial \vz_t} + \epsilon \sqrt{2 * \alpha}, \text{ where } \epsilon \sim \mathcal{N}(0, I_d)
\end{align*}
Next, the proposed $\tilde{z}_{t+1}$ is accepted or rejected using the Metropolis Hastings algorithm, by computing the acceptance ratio:
\begin{align}
r &= \frac{p(\tilde{\z}_{t+1}) q(\z_t | \tilde{\z}_{t+1})}{p(\z_t)q(\tilde{\z}_{t+1} | \z_t)} \\
\frac{p(\tilde{\z}_{t+1})}{p(\z_t)} &= 
\exp\big\{-E_\theta(G_\omega(\tilde{\z}_{t + 1})) + E_\theta(G_\omega(\z_t))\big\} \\
q(\tilde{\z}_{t+1} | \z_t) &\propto \exp\bigg(\frac{-1}{4\alpha} \big|\big|\tilde{\z}_{t+1} - \z_t + \alpha \frac{\partial E_\theta(G_\omega(\z_t))}{\partial \z_t} \big|\big|_2^2\bigg)
\end{align}
and accepting (setting $\vz_{t+1}=\tilde{\vz}_{t+1}$) with probability $r$.

\section{Experiments} 
To understand the benefits of MEG, we first visualize the energy densities learnt by our generative model on toy data. Next, we evaluate the efficacy of our entropy maximizer by running discrete mode collapse experiments to verify that we learn all modes and the corresponding mode count (frequency) distribution. Furthermore, we evaluate the performance of MEG on sharp image generation, since this is a common failure mode of models trained with maximum likelihood which tend to generate blurry samples \citep{theis2015note}.
We also compare MCMC samples in visible space and our proposed sampling from the latent space of the composed energy function. Finally, we run anomaly detection experiments to test the application of the learnt energy function.


We've released \href{https://github.com/ritheshkumar95/energy_based_generative_models}{open-source code}\footnote{\url{https://github.com/ritheshkumar95/energy_based_generative_models}} for all the experiments.

\subsection{Visualizing the learned energy function}
\begin{figure*}[h]
    \centering
    \begin{subfigure}{0.3\textwidth}
        \centering
        \includegraphics[width=.75\linewidth]{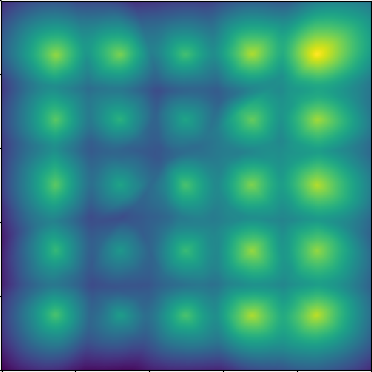}
        \caption{}
    \end{subfigure}
    \begin{subfigure}{0.3\textwidth}
        \centering
        \includegraphics[width=.75\linewidth]{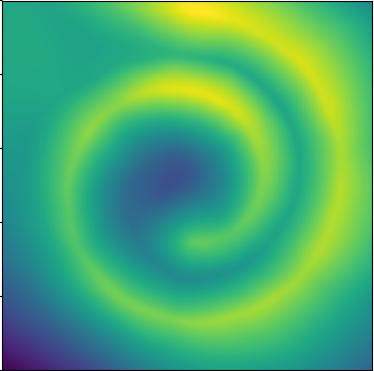}
        \caption{}
    \end{subfigure}
    \begin{subfigure}{0.3\textwidth}
        \centering
        \includegraphics[width=.75\linewidth]{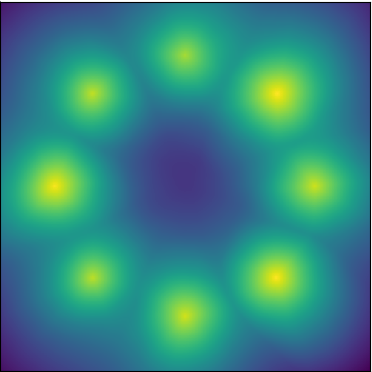}
        \caption{}
    \end{subfigure}
    \caption{Probability density visualizations for three popular toy dataset (a) 25-gaussians, (b) swiss roll, and (c) 8-gaussians. Density was estimated using a sample based approximation of the partition function.}
    \label{fig:mnist}
\end{figure*}

Generative models trained with maximum likelihood often suffer from the problem of spurious modes and excessive entropy of the trained distribution, where the model incorrectly assigns high probability mass to regions not present in the data manifold. Typical energy-based models such as RBMs suffer from this problem partly because of the poor approximation of the negative phase gradient, as discussed above, and the large price paid in terms of log-likelihood for not putting enough probability mass near data points (i.e. for missing modes).

To check if MEG suffers from spurious modes, we train the energy-based model on synthetic 2D datasets (swissroll, 25gaussians and 8gaussians) and visualize the energy function. From the probability density plots on Figure~\ref{fig:mnist}, we can see that the energy model doesn't suffer from spurious modes and learns a sharp distribution.

\subsection{Investigating Mode Collapse}
GANs are notorious for having mode collapse issues wherein certain modes of the data distribution are not represented by the generated distribution.
Since the generator is trained to minimize its KL divergence with the energy model distribution (which is trained via maximum likelihood), we expect the generator to faithfully capture all the modes of the data distribution. Our theory requires we maximize entropy of the generated distribution, which we believe is instrumental in ensuring full mode capture.

To empirically verify MEG captures all the modes of the data distribution, we follow the same experimental setup as \citep{metz2016unrolled} and \citep{srivastava2017veegan}. We train our generative model on the StackedMNIST dataset, which is a synthetic dataset created by stacking MNIST on different channels. The number of modes are counted using a pretrained MNIST classifier, and the KL divergence is calculated empirically between the generated mode distribution and the data distribution.
\begin{table}[H]
	
\begin{center}
    \caption{Number of captured modes and Kullback-Leibler divergence between the training and samples distributions for ALI~\citep{dumoulin2016adversarially}, Unrolled GAN~\citep{metz2016unrolled}, VeeGAN~\citep{srivastava2017veegan},
	WGAN-GP~\citep{gulrajani2017improved}. Numbers except MEG and WGAN-GP are borrowed from \cite{belghazi2018mine}}
  	\begin{tabular}{lcc}
  	        \toprule
  	        (Max $10^3$) & Modes & KL\\
  	        \midrule
    		Unrolled GAN  & $48.7$ & $4.32$\\
    		VEEGAN  		& $150.0$ &  $2.95$\\
			WGAN-GP         & $959.0$	& $0.7276$ \\
			\midrule
			MEG (ours)   & 1000.0 &  0.0313 \\ \bottomrule
  	\end{tabular}
  	\begin{tabular}{lcc}
  	        \toprule
  	        (Max $10^4$) & Modes & KL\\
  	        \midrule
			WGAN-GP  	& 9538.0 & 0.9144 \\ \midrule
			MEG (ours)   & 10000.0 &  0.0480 \\ \bottomrule
  	\end{tabular}
	\end{center}
	
	\label{table:stacked_mnist}
\end{table}

From Table 1, we can see that MEG naturally covers all the modes in that data, without dropping a single mode. Apart from just representing all the modes of the data distribution, MEG also better matches the data distribution as evidenced by the significantly smaller KL divergence score compared to the baseline WGAN-GP.

Apart from the standard 3-StackMNIST, we also evaluate MEG on a new dataset with $10^4$ modes (4 stacks)\footnote{The 4-StackedMNIST was created in a way analogous to the original 3-StackedMNIST dataset. We randomly sample and fix $128 \times 10^4$ images to train the generative model and take $26 \times 10^4$ samples for evaluations.} which is evidence that MEG does not suffer from mode collapse issues unlike state-of-the-art GANs like WGAN-GP.

\subsection{Modeling Natural Images}
While the energy landscapes in Figure~\ref{fig:mnist} provide evidence that MEG trains energy models with sharp distributions, we next investigate if this also holds when learning a distribution over high-dimensional natural images. Energy-based models trained with existing techniques produce blurry samples due to the energy function not learning a sharp distribution.


We train MEG on the standard benchmark 32x32 CIFAR10 dataset for image modeling. We additionally train MEG on the 64x64 cropped CelebA - celebrity faces dataset to report qualitative samples from MEG. Similar to recent GAN works \citep{miyato2018spectral}, we report both Inception Score (IS) and Fréchet Inception Distance (FID) scores on the CIFAR10 dataset and compare it with a competitive WGAN-GP baseline.

\begin{table}[h]
\centering
  \caption{\label{tab:incepscores}Inception scores and FIDs with unsupervised image generation on CIFAR-10. We used 50000 sample estimates to compute Inception Score and FID.}
\small{
  \begin{tabular}[t]{lrr}
    \toprule
    Method & \textbf{Inception score} & \textbf{FID} \\
    \midrule
    Real data & 11.24$\pm$.12 & 7.8 \\
    WGAN-GP & 6.81 $\pm$ .08 & 30.95\\  
    \midrule
    MEG (Generator) & 6.49 $\pm$ .05 & 35.02\\
    MEG (MCMC) & \textbf{7.31 $\pm$ .06} & 33.18\\
    \bottomrule
  \end{tabular}
}
\label{fig-cifar}
\end{table}

From Table~\ref{fig-cifar}, we can see that in addition to learning an energy function, MEG-trained generative model produces samples comparable to recent GAN methods such as WGAN-GP \citep{gulrajani2017improved}. Note that the perceptual quality of the samples improves by using the proposed MCMC sampler in the latent space. See also Figure \ref{mcmc:visible} in Appendix B for an ablation study which shows that MCMC on the visible space does not perform as well as MCMC on the latent space.



\subsection{Anomaly Detection}
Apart from the usefulness of energy estimates for relative density estimation (up to the normalization constant), energy functions can also be useful to perform unsupervised anomaly detection. Unsupervised anomaly detection is a fundamental problem in machine learning, with critical applications in many areas, such as cyber-security, complex system management, medical care, etc. Density estimation is at the core of anomaly detection since anomalies are data points residing in low probability density
areas. We test the efficacy of our energy-based density model for anomaly detection using two popular benchmark datasets: KDDCUP and MNIST. 

\paragraph{KDDCUP} We first test our generative model on the KDDCUP99 10 percent dataset from the UCI repository \citep{lichman2013uci}. Our baseline for this task is Deep Structured Energy-based Model for Anomaly Detection (DSEBM) \citep{zhai2016deep}, which trains deep energy models such as Convolutional and Recurrent EBMs using denoising score matching \citep{vincent2011connection} instead of maximum likelihood, for performing anomaly detection. We also report scores on the state of the art DAGMM \citep{zong2018deep}, which learns a Gaussian Mixture density model (GMM) over a low dimensional latent space produced by a deep autoencoder. We train MEG on the KDD99 data and use the score norm $||\nabla_x E_\theta(x)||_2^2$ as the decision function, similar to \cite{zhai2016deep}.

\begin{table}[h]
\caption{Performance on the KDD99 dataset. Values for OC-SVM, DSEBM
values were obtained from \cite{zong2018deep}. Values for MEG are derived from 5 runs. For each individual run, the metrics are averaged over the last 10 epochs.}
\label{KDD}
\begin{center}
\small{
\begin{tabular}{c|c|c|c}
\hline 
{\bf Model} & Precision & Recall & F1
\\ \hline 
Kernel PCA & 0.8627  & 0.6319  & 0.7352   \\ 
OC-SVM     & 0.7457  & 0.8523  & 0.7954   \\ 
DSEBM-e    & 0.8619  & 0.6446  & 0.7399   \\
DAGMM      & 0.9297  & 0.9442  & 0.9369   \\ \midrule
MEG (ours) & \textbf{0.9354} $\pm$ \textbf{0.016} & \textbf{0.9521} $\pm$ \textbf{0.014} & \textbf{0.9441} $\pm$ \textbf{0.015} \\ \bottomrule
\end{tabular}
}
\end{center}
\label{table-anomaly}
\end{table}

From Table~\ref{table-anomaly}, we can see that the MEG energy function outperforms the previous SOTA energy-based model (DSEBM) by a large margin (+0.1990 F1 score) and is comparable to the current SOTA model (DAGMM) specifically designed for anomaly detection. Note that DAGMM is specially designed for anomaly detection, while MEG is a general-purpose energy-based model.

\paragraph{MNIST} Next we evaluate our generative model on anomaly detection of high dimensional image data. We follow the same experiment setup as \citep{zenati2018efficient} and make each digit class an anomaly and treat the remaining 9 digits as normal examples. We also use the area under the precision-recall curve (AUPRC) as the metric to compare models.
\begin{table}[h]
\caption{Performance on the unsupervised anomaly detection task on MNIST measured by area under precision recall curve. Numbers except ours are obtained from \citep{zenati2018efficient}. Results for MEG are averaged over the last 10 epochs to account for the variance in scores.}
\label{anomaly_mnist}
\begin{center}
\begin{tabular}{c|c|c|c}
\toprule
{\bf Heldout Digit} & VAE & MEG & BiGAN-$\sigma$
\\ \hline 
1 & 0.063  & 0.281 $\pm$ 0.035 & 0.287 $\pm$ 0.023 \\ 
4 & 0.337  & 0.401 $\pm$ 0.061 & 0.443 $\pm$ 0.029 \\ 
5 & 0.325  & 0.402 $\pm$ 0.062 & 0.514 $\pm$ 0.029 \\ 
7 & 0.148  & 0.29  $\pm$ 0.040 & 0.347 $\pm$ 0.017 \\ 
9 & 0.104  & 0.342 $\pm$ 0.034 & 0.307 $\pm$ 0.028 \\
\bottomrule
\end{tabular}

\vspace{-2em}
\end{center}
\end{table}
From Table 4, it can be seen that our energy model outperforms VAEs for outlier detection and is comparable to the SOTA BiGAN-based anomaly detection methods for this dataset \citep{zenati2018efficient} which train bidirectional GANs to learn both an encoder and decoder (generator) simultaneously and use a combination of the reconstruction error in output space as well as the discriminator's cross entropy loss as the decision function.

Our aim here is not to claim state-of-the-art on the task of anomaly detection but to demonstrate the quality of the energy functions learned by our technique, as judged by its competitive performance on anomaly detection.

\section{Related Work}

Early work on deep learning relied on unsupervised learning \citep{Hinton06,Bengio-nips-2006,Larochelle-jmlr-2009} to train energy-based models~\citep{lecun2006}, in particular Restricted Boltzmann Machines, or RBMs. \citet{Hinton-PoE-2000} proposed $k$-step Contrastive Divergence (CD-$k$), to efficiently approximate the negative phase log-likelihood gradient. Subsequent work have improved on CD-$k$ such as Persistent CD \citep{salakhutdinov2009deep, tieleman2008training}. \citet{hyvarinen2005estimation} proposed an alternative method to train non-normalized graphical models using Score Matching, which does not require computation of the partition function.

\citet{Kim+Bengio-2016} and \citet{Dai-et-al-2017} also learn a generator that approximates samples from an energy-based model. However, their approach for entropy maximization is different from our own. \citet{Kim+Bengio-2016} argue that batch normalization \citep{ioffe2015batch} makes the hidden activations of the generator network approximately Gaussian distributed and thus maximize the log-variance for each hidden activation of the network. \citet{Dai-et-al-2017} propose two approaches to entropy maximization. One which minimizes entropy of the inverse model ($p_{gen}(z|x)$) which is approximated using an amortized inverse model similar to ALI \citep{dumoulin2016adversarially}, and another which makes isotropic Gaussian assumptions for the data.
In our work, we perform entropy maximization using a tight mutual information estimator which does not make any assumptions about the data distribution.


\cite{Zhao-et-al-2016} use an autoencoder as the discriminator and use the reconstruction loss as a signal to classify between real and fake samples. The autoencoder is highly regularized to allow its interpretation as an energy function. However \cite{Dai-et-al-2017} prove that the EBGAN objective does not guarantee the discriminator to recover the true energy function. The generator diverges from the true data distribution after matching it, since it would continue to receive training signal from the discriminator. The discriminator signal does not vanish even at optimality (when $P_G = P_D$) if it retains density information, since some samples would be considered "more real" than others.

\vspace{-1em}
\section{Conclusion}
\vspace{-1em}
We proposed MEG, an energy-based generative model that produces energy estimates using an energy model and a generator that produces fast approximate samples. This takes advantage of novel methods to maximize the entropy at the output of the generator using a nonparametric mutual information lower bound estimator. We have shown that our energy model learns good energy estimates using visualizations in toy 2D datasets and through performance in unsupervised anomaly detection. We have also shown that our generator produces samples of high perceptual quality by measuring Inception Scores and Fréchet Inception Distance and shown that MEG is robust to the respective weaknesses of GAN models (mode dropping) and maximum-likelihood energy-based models (spurious modes).

\section{Acknowledgements}
The authors acknowledge the funding provided by NSERC,
Canada AI Chairs, Samsung, Microsoft, Google and Facebook. We also thank NVIDIA for donating a DGX-1 computer used for certain experiments in this work.
\bibliographystyle{neurips_2019}
\bibliography{neurips_2019}

\newpage
\begin{appendices}
\section{Training Algorithm}
\begin{algorithm}[h]
\caption{\small \,\,\textbf{MEG Training Procedure} Default values: Adam parameters $\alpha = 0.0001, \beta_1 = 0.5, \beta_2 = 0.9; \lambda = 0.1$; $n_{\varphi} = 5$}
\small{
\begin{algorithmic}
\REQUIRE{Score penalty coefficient $\lambda$, $\#$ of $\theta$ updates per generator update $n_\varphi$ , $\#$ of training iterations $T$}, Adam hyperparameters $\alpha$, $\beta_1$ and $\beta_2$.
\REQUIRE{Energy function $E_\theta$ with parameters $\theta$, entropy statistics function $T_\phi$ with parameters $\phi$, generator function $G_\omega$} with parameters $\omega$, minibatch size $m$,
\FOR{$t = 1, ... , T$}
 \FOR{$1, ..., n_{\varphi}$}
	\STATE{Sample minibatch of real data $\{\x^{(1)},...,\x^{(m)}\} \sim P_D$.}
	\STATE{Sample minibatch of latent $\{\z^{(1)}_0,...,\z^{(m)}_0\} \sim P_z$.}
	\STATE{$\tilde{\x} \leftarrow G_\omega(\z)$}
	\STATE{\vspace{1mm}
	    $\mathcal{L}_E \leftarrow \frac{1}{m} \bigg[\sum_i^m E_\theta(\x^{(i)})  - \sum_i^m E_\theta(\tilde{\x}^{(i)}) + \lambda \sum_i^m \big|\big|\nabla_{\x^{(i)}} E_\theta(\x^{(i)})\big|\big|^2 \bigg]$
    }
    \STATE{$\theta \leftarrow \text{Adam}(\displaystyle \mathcal{L}_E, \theta, \alpha, \beta_1, \beta_2$) } 
  \ENDFOR
	\STATE{Sample minibatch of latent $\z=\{\z^{(1)},...,\z^{(m)}\} \sim P_z$.}
	\STATE{Per-dimension shuffle of $\z$, yielding $\{\tilde{\z}^{(1)},...,\tilde{\z}^{(m)}\}$.}
	\STATE{$\tilde{\x} \leftarrow G_\omega(\z)$}
    \STATE{$\displaystyle
    \mathcal{L}_H \leftarrow  \frac{1}{m}\sum_i^m \bigg[ \log \sigma( T_\phi(\tilde{\x}^{(i)}, \z^{(i)})) - \log \big(1 - \sigma( T_\phi(\tilde{\x}^{(i)}, \tilde{\z}^{(i)}))\big) \bigg]
    $}
    \STATE{$\displaystyle
    \mathcal{L}_G \leftarrow  \frac{1}{m} \bigg[\sum_i^m E_\theta(\tilde{\x}^{(i)})\bigg] + \mathcal{L}_H
    $}
    \STATE{$\omega \leftarrow \text{Adam}(\displaystyle \mathcal{L}_G, \omega, \alpha, \beta_1, \beta_2$)}
    \STATE{$\phi \leftarrow \text{Adam}(\displaystyle \mathcal{L}_H, \phi, \alpha, \beta_1, \beta_2$)}
\ENDFOR
\end{algorithmic}
}
\label{alg:MEG}
\end{algorithm}

\pagebreak
\section{MCMC sampling in visible vs latent space}
\begin{figure*}[h]
\centering
\begin{subfigure}[t]{0.4\textwidth}
\includegraphics[width=1\linewidth]{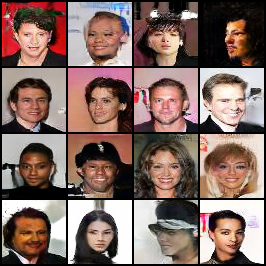}
\end{subfigure}
\begin{subfigure}[t]{0.4\textwidth}
\includegraphics[width=1\linewidth]{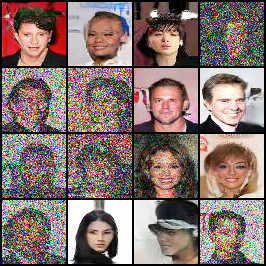}
\end{subfigure}
\\
\begin{subfigure}[b]{.4\textwidth}
\includegraphics[width=1\linewidth]{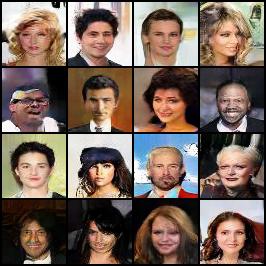}
\end{subfigure}
\begin{subfigure}[b]{.4\textwidth}
\includegraphics[width=1\linewidth]{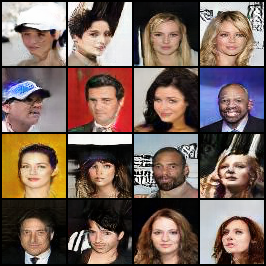}
\end{subfigure}

\caption{Samples from the beginning and end of the MCMC in visible space (\textbf{top}) and latent space {\textbf{bottom}} using the MALA proposal and acceptance criteria. MCMC in visible space has poor mixing and gets attracted to spurious modes, while MCMC in latent space seems to change semantic attributes of the image, while not producing spurious modes. }
\label{mcmc:visible}
\end{figure*}




\end{appendices}

\end{document}